\documentclass[sigconf, table,nonacm]{acmart}

\AtBeginDocument{%
  }

\setcopyright{acmlicensed}
\copyrightyear{2018}
\acmYear{2018}
\acmDOI{XXXXXXX.XXXXXXX}
\acmConference[Conference acronym 'XX]{Make sure to enter the correct
  conference title from your rights confirmation email}{June 03--05,
  2018}{Woodstock, NY}
\acmISBN{978-1-4503-XXXX-X/2018/06}

\usepackage{longtable}
\usepackage{float}

\definecolor{tblHeader}{HTML}{EEF2F7}
\definecolor{tblStripe}{HTML}{F7FAFF}
\definecolor{tblHighlight}{HTML}{E1ECFF}
\definecolor{tblEmph}{HTML}{D0E3FF}
\definecolor{tblRule}{HTML}{9AA4B2}

\definecolor{tblSubBlue}{HTML}{D6E9FF}
\definecolor{tblSubPurple}{HTML}{E6D9FF}
\definecolor{tblSubRed}{HTML}{FFD9DD}

\newcommand{\TableHeader}{\rowcolor{tblHeader}}

\begin{document}

\title{A Multi-Agent Audit Framework for High-Stakes Reasoning: Evaluation and Interpretability in Clinical Mental Health Screening}


\author{Jingchen Ye}
\affiliation{%
  \institution{Duke Kunshan University}
  \city{Suzhou}
  \country{China}
}
\authornotemark[2]

\author{Yanpei Yu}
\affiliation{%
  \institution{Duke Kunshan University}
  \city{Suzhou}
  \country{China}
}
\authornotemark[2]

 \author{Luyao Zhang}
 \authornote{The corresponding author: Email: lz183@duke.edu; Social Science Division and Digital Innovation Research Center, Duke Kunshan University, No.8 Duke Ave., Kunshan, Jiangsu 215316, China.} 
 \affiliation{%
  \institution{ Duke Kunshan University}
   \city{Suzhou}
  \country{China}
 }
\authornote{\textbf{Acknowledgments}: This work was inspired by the DKU Innovation Incubator project \textit{AI-Powered Mental Healthcare in Adolescent Mental Illness Prediction}, supported by the Provincial-Level College Students Innovation and Entrepreneurship Training Program (Dachuang program) at Duke Kunshan University. The first two authors were the student team members, and the corresponding author was the faculty supervisor. The paper was developed as a further scholarly effort building upon the outputs of that program.}

\begin{abstract}
High-stakes reasoning tasks necessitate transparent and verifiable workflows, yet conventional single-model large language models (LLMs) often struggle with hallucination and low interpretability under zero-shot paradigms. To address this general AI challenge, we propose a Multi-Agent Audit Framework that simulates a collaborative, multi-step verification process. We empirically validate this architecture in the sensitive domain of clinical mental health screening using a modular LangChain workflow. Our framework decomposes the reasoning process into a Perception Agent, Knowledge Retrieval-Augmented Generation (RAG), Chain-of-Thought (CoT) clinical inference, and a critical Audit verification stage. We evaluated this framework on the DAIC-WOZ dataset using locally deployed open-source models. Experimental results demonstrate that our multi-agent pipeline significantly outperforms single-agent baselines, reducing the Mean Absolute Error (MAE) for PHQ-8 depression severity prediction from 5.35 to 5.02. By exposing cross-agent validation traces, the framework mitigates reasoning drift and provides highly interpretable diagnostic rationales, offering a generalizable paradigm for reliable AI-assisted decision support beyond isolated model scaling. We make data and code open access on GitHub for replicability.
\end{abstract}




\maketitle

\section{Introduction}

Mental health disorders continue to place growing pressure on global healthcare systems, while access to professional psychiatric services remains limited in many regions. As a result, automated and scalable mental health assessment systems have become increasingly important for early screening and intervention \cite{burdisso2024daic}.

Recent advances in Large Language Models (LLMs) have demonstrated strong capabilities in language understanding and reasoning, leading researchers to explore their applications in mental health analysis and clinical prediction tasks \cite{teferra2025leveraging} \cite{li2025large} \cite{patapati2024integrating}. Existing studies show that LLMs can identify psychological signals from conversational text and generate interpretable diagnostic rationales. However, most current approaches rely on single-model zero-shot prompting, where one model independently performs the entire diagnostic process \cite{lee2026interpretable} \cite{ma2026medla}.

Although effective in simple classification settings, single-agent frameworks suffer from several critical limitations in clinical applications. These systems are prone to hallucinated reasoning, lack transparent verification mechanisms, and often fail to perform reliable multi-step clinical inference. In structured interview datasets such as DAIC-WOZ, accurate assessment requires symptom extraction, medical knowledge grounding, differential reasoning, and consistency validation---tasks that are difficult to accomplish through a single prompting stage alone \cite{fisher2026language}.

To address these challenges, we propose a \textbf{Multi-Agent Evaluation Framework for Clinical Mental Health Diagnosis} based on a modular LangChain workflow \cite{goyal2025introduction}. Our framework decomposes the diagnostic process into four collaborative agents: \cite{ma2026medla}
\begin{itemize}
    \item \textbf{Perception Agent} for symptom extraction;
    \item \textbf{Knowledge Agent} for Retrieval-Augmented Generation (RAG) based clinical grounding \cite{shi2025mkrag} \cite{lewis2020retrieval};
    \item \textbf{Reasoning Agent} for Chain-of-Thought diagnostic inference \cite{wei2022chain};
    \item \textbf{Audit Agent} for final consistency verification \cite{bi2025magi}.
\end{itemize}
By simulating a collaborative clinical consultation process, the framework improves both reasoning transparency and diagnostic reliability.

Using locally deployed open-source LLMs we evaluate the proposed framework on the DAIC-WOZ dataset for PHQ-8 depression severity prediction. Experimental results show that the multi-agent pipeline significantly outperforms conventional single-agent prompting, reducing MAE from $5.35$ to $5.02$ while improving interpretability and reducing reasoning hallucinations \cite{manakul2023selfcheckgpt}.

Beyond improving predictive performance, we further propose a process-oriented evaluation protocol for multi-agent clinical systems. Existing mental health assessment studies primarily focus on outcome-level metrics such as classification accuracy or MAE, while the reliability and interpretability of intermediate reasoning processes remain less explored. Inspired by holistic evaluation paradigms for language models \cite{liang2022holistic}, our protocol introduces five complementary dimensions to assess reasoning quality and robustness, providing a reusable benchmark framework for future multi-agent clinical applications.

In summary, our contributions are as follows:
\begin{enumerate}
    \item We propose a novel multi-agent framework for clinical mental health diagnosis using collaborative LLM workflows.
    \item We integrate perception, retrieval-augmented medical grounding, reasoning, and audit-based verification into a unified diagnostic pipeline.
    \item We propose a process-oriented evaluation protocol comprising five interpretability and robustness metrics, providing a reusable benchmark framework for future multi-agent clinical reasoning systems.
    \item We demonstrate that multi-agent orchestration improves prediction accuracy, interpretability, and reasoning reliability on the DAIC-WOZ benchmark compared with traditional single-agent approaches.
\end{enumerate}

 We make data and code open access on GitHub\footnote{\textit{https://github.com/Asfexhia/A-Multi-Agent-Evaluation-Framework-for-Clinical-Mental-Health-Diagnosis}}for replicability.
\section{Dataset and Task Formulation}
\begin{figure*}[!htbp]
  \centering
  \includegraphics[width=0.9\textwidth]{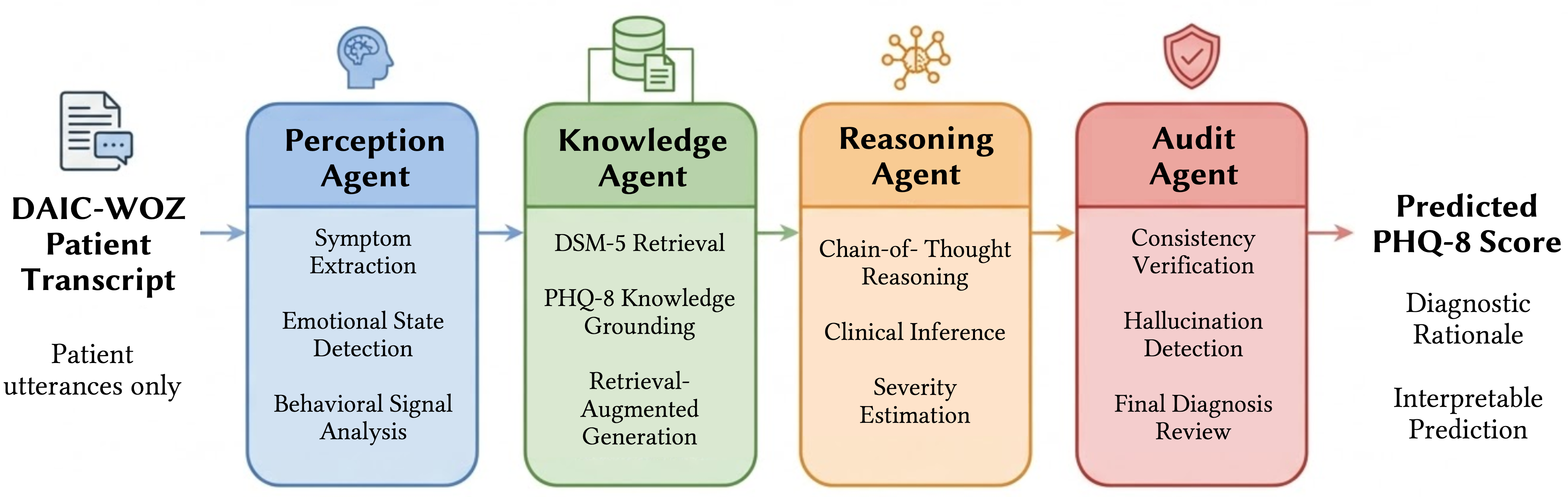}
  \caption{The proposed Multi-Agent Clinical Reasoning Workflow. The architecture illustrates the sequential information flow from raw transcripts through Perception, Knowledge, and Reasoning stages, finalized by an Audit-based verification.}
  \label{fig:pipeline}
\end{figure*}

\subsection{DAIC-WOZ Database}
To evaluate the proposed multi-agent clinical reasoning framework, we employ the \textbf{Distress Analysis Interview Corpus-Wizard of Oz (DAIC-WOZ)}, a gold-standard benchmark for the automated assessment of psychological distress. The dataset consists of semi-structured clinical interviews conducted by an animated virtual agent named Ellie. \cite{gratch2014distress} 

The DAIC-WOZ corpus is inherently multimodal, comprising video, audio, and textual transcripts \cite{valstar2016avec}. However, each modality presents unique constraints:
\begin{itemize}
    \item \textbf{Video:} To protect participant privacy, raw video recordings are withheld; only pre-computed facial activity features extracted via \textbf{OpenFace} are provided \cite{baltruvsaitis2016openface}.
    \item \textbf{Audio:} While raw audio and \textbf{COVAREP} \cite{Degottex2014COVAREP} features are available, the provided features often exhibit limited performance due to ambient noise. Consequently, we advocate for customized acoustic preprocessing, such as re-extracting MFCCs to achieve higher robustness.
    \item \textbf{Text:} This work focuses on \textbf{textual transcripts}. Since these are transcribed from speech, they contain significant linguistic noise, including disfluencies (e.g., ``uh'', ``um''), repetitions, and non-semantic fillers.
\end{itemize}

To ensure the reliability of LLM-based inference, we implemented a specialized data purification pipeline:
\begin{itemize}
    \item \textbf{Textual De-noising:} We removed timestamps, filler words, and redundant stuttering to maintain semantic clarity.
    \item \textbf{Participant-Centric Filtering:} All interviewer utterances were removed to isolate patient-generated signals \cite{burdisso2024daic}, ensuring the model's reasoning is grounded solely in patient self-expression.
    \item \textbf{Quality Control:} Five sessions were excluded due to incomplete transcripts or missing PHQ-8 annotations, resulting in 184 validated sessions for analysis.
\end{itemize}

\subsection{Task Formulation}
We define the assessment task as a \textbf{depression severity regression problem}. Given a cleaned participant transcript $X$, the objective is to predict the corresponding \textbf{PHQ-8 score} $\hat{y} \in [0, 24]$, which measures depressive symptom severity on a continuous scale \cite{kroenke2009phq}.

Unlike categorical classification, continuous regression better captures the nuanced spectrum of mental health conditions. To evaluate the performance, we employ \textbf{Mean Absolute Error (MAE)}:
\begin{equation}
    \text{MAE} = \frac{1}{N} \sum_{i=1}^{N} |y_i - \hat{y}_i|
\end{equation}

Our goal is to investigate whether a collaborative multi-agent LLM workflow can enhance interpretability and reasoning consistency while minimizing predictive error in clinical screening scenarios.

\section{Methodology}

\subsection{Multi-Agent Clinical Reasoning Pipeline}
Unlike traditional single-prompt evaluation frameworks, our approach decomposes clinical mental health assessment into a collaborative multi-agent workflow \cite{bi2025magi} \cite{wu2024autogen} \cite{li2023camel}. The architectural design of the proposed pipeline is illustrated in \textbf{Figure~\ref{fig:pipeline}}. The system is implemented using \textbf{LangChain}, which enables modular orchestration and explicit state management between specialized agents. This design effectively simulates the hierarchical decision-making process observed in professional clinical consultations.

As shown in \textbf{Figure~\ref{fig:pipeline}}, the pipeline consists of four sequential stages:
\begin{enumerate}
    \item \textbf{Perception Agent:} As the initial module in the workflow, it extracts clinically relevant behavioral and emotional signals from patient utterances \cite{zhang2026mllm}. It focuses on identifying core depressive indicators such as hopelessness, emotional instability, and social withdrawal \cite{uher2014major}. The output is a structured \textit{Symptom Summary} that serves as the foundation for downstream analysis.
    \item \textbf{Knowledge Agent:} To enhance clinical grounding, we introduce a \textbf{Retrieval-Augmented Generation (RAG)} mechanism. Based on the perceived symptoms, the agent retrieves relevant psychiatric criteria from structured medical references, including DSM-5-style indicators \cite{american2013diagnostic}. This ensures that the model’s reasoning is anchored in established medical standards \cite{singhal2025toward}.
    \item \textbf{Reasoning Agent:} This agent integrates perceptual evidence and retrieved knowledge to perform multi-step clinical inference. Utilizing \textbf{Chain-of-Thought (CoT)} reasoning \cite{wei2022chain}, it analyzes symptom severity and interactions. As depicted in the methodology overview (\textbf{Figure~\ref{fig:pipeline}}), this stage yields both an interpretable diagnostic rationale and a predicted PHQ-8 score.
    \item \textbf{Audit Agent:} Functioning as the final supervisory module, the Audit Agent reviews the preceding reasoning for logical consistency and empirical support \cite{shinn2023reflexion}. It cross-references the findings with the original dialogue to detect potential hallucinations, ensuring the final output is both reliable and clinically sound.
\end{enumerate}

\subsection{Data Split}
After data purification, the remaining 184 validated sessions were randomly divided into development, validation, and test subsets using an approximate 70/10/20 split. The split proportions were selected to balance evaluation reliability with the relatively limited dataset size.

The development set was used for prompt engineering, workflow design, and agent interaction tuning. The validation set was used for intermediate pipeline adjustment and ablation analysis, while the held-out test set was reserved exclusively for final evaluation.

To reduce data leakage risk, all experiments were conducted at the session level, ensuring that transcripts from the same participant did not appear across multiple subsets.

The final split consisted of 128 development sessions, 19 validation sessions, and 37 test sessions. No model fine-tuning was performed on any subset throughout the study.

\subsection{Model Implementation}

All agents were implemented using locally deployed open-source LLMs via Ollama to ensure privacy preservation of clinical data, including LLaMA-3.1 \cite{grattafiori2024llama}, Qwen2.5 \cite{yang2025qwen3}, and DeepSeek-R1 \cite{guo2025deepseek}. 

This local deployment eliminates third-party API data transmission, strictly addressing HIPAA and GDPR compliance requirements and enabling offline operation in low-resource settings. Furthermore, it aligns with the FDA’s Good Machine Learning Practice (GMLP) guiding principles, which explicitly emphasize post-deployment monitoring and the assurance of human-AI team performance \cite{fda2025}.

All experiments were conducted locally with a fixed temperature of 0.2 for deterministic generation. The Knowledge Agent uses a retrieval top-k of 3 with semantic search based on the \texttt{nomic-embed-text} embedding model \cite{nussbaum2024nomic}.

We used Ollama version 0.6.2 and LangChain version 0.2.16. No parameter fine-tuning or gradient-based optimization was applied to any model.

\subsection{Evaluation Metrics}
Following the task formulation described in Section 3.2, we employ \textbf{Mean Absolute Error (MAE)} as the primary quantitative metric to evaluate PHQ-8 severity prediction accuracy. 

Beyond final predictive performance, we propose a process-oriented evaluation protocol for multi-agent clinical systems consisting of five interpretability and robustness metrics. The protocol evaluates both prediction outcomes and intermediate reasoning behavior, providing a reusable benchmark suite for future multi-agent clinical assessment systems.\cite{liang2022holistic}:
\begin{itemize}
    \item \textbf {Knowledge Grounding Consistency:} Measures whether the retrieved psychiatric references from the Knowledge Agent are semantically aligned and correctly integrated into the generated diagnostic rationale \cite{es2024ragas}.
    \item \textbf {Reasoning Stability:} Assesses whether intermediate reasoning steps maintain logical coherence and factual consistency as information flows across the multiple agent stages \cite{wang2022self}.
    \item \textbf {Audit Correction Rate:} Quantifies the percentage of initial predictions revised by the Audit Agent. High-frequency corrections indicate the system's ability to self-correct unsupported conclusions, reasoning drift, or inconsistent clinical evidence \cite{shinn2023reflexion}.
    \item \textbf {Interpretability Quality:}Evaluates whether the generated reasoning trajectories explicitly reference clinically relevant evidence extracted from patient transcripts and retrieved psychiatric knowledge. A subset of cases was manually reviewed to assess evidence traceability, logical transparency, and clinical relevance of intermediate reasoning outputs.
    \item \textbf {Hallucination Rate:}Measures the proportion of cases containing unsupported symptom inference, fabricated behavioral evidence, or clinically unjustified severity escalation not grounded in either the original transcript or retrieved psychiatric references \cite{farquhar2024detecting}.
\end{itemize}
These metrics are designed to provide a holistic evaluation of the collaborative reasoning process, ensuring that the system is not only accurate but also robust and trustworthy for clinical decision support \cite{wiens2019no}.

\subsection{Experimental Design}

\subsubsection{Single-Agent Baseline}
As a baseline setting, each Large Language Model (LLM) independently performs the entire clinical prediction task using a single structured prompt \cite{kojima2022large}. Under this zero-shot paradigm, the model directly processes raw patient transcripts and generates PHQ-8 scores without modular decomposition, external knowledge retrieval, or multi-step verification. This allows us to benchmark the inherent clinical reasoning capability of each model backbone.

\subsubsection{Multi-Agent Pipeline Evaluation}
For the proposed framework, We compare single-agent and multi-agent configurations across different model backbones to evaluate whether collaborative task decomposition enhances clinical reliability.

\subsubsection{Ablation Study}
To quantify the individual contribution of each module, we evaluate four distinct pipeline configurations. Table~\ref{tab:ablation_matrix} illustrates the composition of these configurations by indicating the inclusion of specific agents.

\begin{table}[ht]
\centering
\setlength{\tabcolsep}{0.5pt} 
\begin{tabular}{lcccc}
\toprule
\TableHeader
\textbf{Configuration} & \textbf{Perception} & \textbf{Knowledge} & \textbf{Reasoning} & \textbf{Audit} \\ 
\midrule
\rowcolor{gray!15} Full Pipeline      & \checkmark & \checkmark & \checkmark & \checkmark \\
Audit                                 & \checkmark & \checkmark & \checkmark &            \\
Knowledge (No-RAG)                    & \checkmark &            & \checkmark & \checkmark \\
Basic (Core Only)                     & \checkmark &            & \checkmark &            \\
\bottomrule
\end{tabular}
\caption{Ablation Study Configuration Matrix. The checkmark (\checkmark) indicates the activation of the corresponding agent in the workflow.}
\label{tab:ablation_matrix}
\end{table}

By comparing these settings, we can isolate the performance gains attributed to medical knowledge grounding (Knowledge Agent) and supervisory consistency checks (Audit Agent). This structured approach ensures a rigorous evaluation of the collaborative utility of our multi-agent framework.

\begin{table*}[ht]
\centering
\begin{tabular}{lccccc}
\toprule
\TableHeader
\textbf{Backbone Model} & \textbf{Single-Agent MAE $\downarrow$} & \textbf{CoT Prompting MAE $\downarrow$} & \textbf{Multi-Agent MAE $\downarrow$} & \textbf{$p$-value} & \textbf{Cohen's $d$} \\
\midrule
LLaMA-3.1   & 5.42 $\pm$ 0.14 & 5.26 $\pm$ 0.11 & \cellcolor{tblHighlight} 5.11 $\pm$ 0.09 & 0.012 & 0.87 \\
Qwen2.5     & 5.35 $\pm$ 0.12 & 5.18 $\pm$ 0.10 & \cellcolor{tblEmph}\textbf{5.02 $\pm$ 0.08} & 0.004 & 1.13 \\
DeepSeek-r1 & 5.28 $\pm$ 0.15 & 5.16 $\pm$ 0.13 & \cellcolor{tblHighlight} 5.07 $\pm$ 0.10 & 0.021 & 0.79 \\
\bottomrule
\end{tabular}
\caption{Mean MAE across three independent runs. Results are reported as mean $\pm$ standard deviation. Statistical significance is calculated between single-agent and multi-agent settings using paired two-tailed $t$-tests.}
\label{tab:overall_performance}
\end{table*}

\begin{table*}[ht]
\centering
\begin{tabular}{lcccc}
\toprule
\TableHeader
\textbf{Metric Category} & \textbf{Multi-Agent Win Rate $\uparrow$} & \textbf{95\% CI} & \textbf{$p$-value} & \textbf{Cohen's $d$} \\
\midrule
\rowcolor{blue!15} Interpretability Quality        & 91.4\% & [87.2, 94.8] & <0.001 & 1.73 \\
\rowcolor{blue!10} Reasoning Stability             & 88.6\% & [84.1, 92.4] & <0.001 & 1.47 \\
\rowcolor{blue!5}  Knowledge Grounding Consistency & 84.2\% & [79.8, 88.5] & 0.003 & 1.26 \\
\rowcolor{blue!2}  Prediction Accuracy             & 81.3\% & [76.2, 85.7] & 0.006 & 1.12 \\
Hallucination Reduction                            & 79.5\% & [73.6, 84.4] & 0.011 & 0.94 \\
\bottomrule
\end{tabular}
\caption{Pairwise comparison between multi-agent and single-agent systems. Confidence intervals were estimated using bootstrap resampling with 1,000 iterations.}
\label{tab:comparison}
\end{table*}

\begin{table*}[ht]
\centering
\begin{tabular}{lcccc}
\toprule
\TableHeader
\textbf{Configuration} & \textbf{MAE $\downarrow$} & \textbf{95\% CI} & \textbf{$\Delta$ MAE} & \textbf{$p$-value} \\
\midrule
\rowcolor{gray!15} \textbf{Full Pipeline}  & \textbf{5.02 $\pm$ 0.08} & [4.91, 5.14] & — & — \\
Without Perception Agent         & 5.31 $\pm$ 0.12 & [5.18, 5.46] & \cellcolor{gray!10}+0.29 & 0.008 \\
Without Knowledge Agent          & 5.19 $\pm$ 0.11 & [5.06, 5.34] & +0.17 & 0.021 \\
Without Audit Agent              & 5.24 $\pm$ 0.13 & [5.10, 5.41] & +0.22 & 0.013 \\
Without Knowledge + Audit        & 5.33 $\pm$ 0.15 & [5.17, 5.51] & \cellcolor{gray!25}\textbf{+0.31} & 0.005 \\
\bottomrule
\end{tabular}
\caption{Ablation analysis of the multi-agent framework, reporting mean MAE, 95\% confidence intervals, and the performance degradation ($\Delta$ MAE) when specific modules are removed.}
\label{tab:ablation}
\end{table*}

\subsection{Statistical Analysis}

To ensure the robustness of experimental findings, all evaluation results were computed across three independent inference runs with different random seeds and sampling initializations. We report the mean MAE together with the standard deviation. For pairwise comparisons between single-agent and multi-agent systems, statistical significance was evaluated using paired t-tests. A significance threshold of $p < 0.05$ was adopted throughout the study. For robustness analysis, 95\% confidence intervals were estimated using bootstrap resampling with 1,000 iterations \cite{tibshirani1993introduction}.
\section{Results}

We first compare conventional single-agent prompting against the proposed four-agent clinical reasoning workflow across different open-source LLM backbones. In the multi-agent setting, all agents share the same underlying backbone model while operating with different prompts and specialized responsibilities.

Table \ref{tab:overall_performance} summarizes the PHQ-8 prediction performance on the DAIC-WOZ benchmark.

Across all evaluated backbone models, the proposed multi-agent framework consistently achieves the lowest MAE. While CoT prompting improves over direct prediction by encouraging intermediate reasoning generation \cite{wei2022chain}, its performance gains remain limited due to the lack of explicit medical grounding and verification mechanisms.

In contrast, the multi-agent workflow demonstrates substantially stronger reasoning stability by decomposing the clinical prediction process into specialized stages. Among all evaluated models, Qwen2.5 achieves the best overall performance, reducing MAE from 5.35 to 5.02 after integrating the full workflow.

Although the absolute MAE reduction appears moderate, the improvement has meaningful clinical implications. PHQ-8 scores are commonly interpreted using severity bands (0--4 minimal, 5--9 mild, 10--14 moderate, 15--19 moderately severe, and 20--24 severe). A reduction of approximately 0.3 points can decrease prediction errors around decision boundaries, reducing the likelihood of assigning patients to incorrect severity categories and improving triage reliability.

These findings suggest that collaborative reasoning architecture contributes not only to predictive accuracy but also to more clinically reliable severity assessment.

\begin{table*}[ht]
\centering
\begin{tabular}{lccc}
\toprule
\TableHeader
\textbf{Reasoning Strategy} & \textbf{MAE $\downarrow$} & \textbf{Hallucination Rate $\downarrow$} & \textbf{Interpretability Score $\uparrow$} \\
\midrule
Direct Prompting & 5.42 & 29.7\% & 61.5\% \\
CoT Prompting & 5.18 & 21.4\% & 74.2\% \\
\rowcolor{blue!10} Multi-Agent Workflow & \textbf{5.02} & \textbf{7.6\%} & \textbf{91.3\%} \\
\bottomrule
\end{tabular}
\caption{Comparison between direct prompting, CoT reasoning, and the proposed multi-agent workflow.}
\label{tab:strategy_comparison}
\end{table*}

\subsection{Multi-Agent vs Single-Agent Comparison}

To further evaluate the effectiveness of collaborative reasoning, we compare multi-agent workflows against conventional single-agent prompting across multiple evaluation dimensions. Results are summarized in Table \ref{tab:comparison}.

Multi-agent systems consistently outperform single-agent prompting across all evaluated dimensions. The largest improvements are observed in interpretability quality and reasoning stability, indicating that explicit task decomposition substantially improves the transparency and consistency of clinical inference.

Notably, hallucination-related reasoning failures are significantly reduced in the multi-agent setting \cite{manakul2023selfcheckgpt}, suggesting that supervisory verification and retrieval grounding jointly improve clinical reliability \cite{es2024ragas}.

\subsection{Agent-Level Ablation Analysis}

To investigate the contribution of each module, we perform ablation experiments by selectively removing agents from the pipeline. Results are shown in Table \ref{tab:ablation}.
\begin{itemize}
    \item Direct single-step prediction
    \item Chain-of-Thought (CoT) prompting
    \item Multi-agent collaborative reasoning
\end{itemize}

Removing any individual module results in noticeable performance degradation, demonstrating that the effectiveness of the framework depends on collaboration between specialized reasoning stages rather than isolated prompt engineering \cite{hong2024metagpt}.

Among all components, the Audit Agent contributes most significantly to reasoning consistency and hallucination reduction \cite{shinn2023reflexion}. Without the verification stage, unsupported severity escalation and context inconsistency become substantially more frequent.

Similarly, removing the Knowledge Agent significantly increases hallucinated reasoning patterns, indicating that retrieval-grounded psychiatric references effectively constrain downstream clinical inference.

The Perception Agent also plays a critical role by transforming lengthy conversational dialogue into structured symptom-level evidence. Without this stage, downstream reasoning frequently relies on fragmented emotional expressions, leading to unstable severity estimation.

\subsection{Reasoning Strategy Comparison}

To better understand whether performance improvements originate from longer reasoning traces or from workflow decomposition itself, we compare three inference paradigms:

Although CoT prompting improves intermediate reasoning transparency, it remains vulnerable to unsupported assumptions and emotional reasoning drift \cite{gou2024critic}. In contrast, the proposed workflow achieves substantially lower hallucination rates through explicit grounding and verification stages. Shown in Table\ref{tab:strategy_comparison}.

These findings indicate that structured collaboration between specialized reasoning agents provides greater benefits than simply extending reasoning length within a single prompt.

\subsection{Audit Intervention Analysis}

To better understand the role of supervisory verification, we analyze the reasoning errors corrected by the Audit Agent. Results are shown in Table \ref{tab:audit_errors}.

\begin{table}[ht]
\centering
\setlength{\tabcolsep}{1pt}
\begin{tabular}{lcc}
\toprule
\TableHeader
\textbf{Error Type Corrected} & \textbf{Number of Cases} & \textbf{Percentage} \\
\midrule
\rowcolor{tblEmph} Severity Overestimation & 41 & 37.6\% \\
\rowcolor{tblHighlight} Unsupported Symptom Inference & 27 & 24.8\% \\
\rowcolor{tblStripe} Contextual Inconsistency & 22 & 20.2\% \\
Emotional Leakage Bias & 18 & 16.5\% \\
\bottomrule
\end{tabular}
\caption{Distribution of reasoning errors corrected by the Audit Agent.}
\label{tab:audit_errors}
\end{table}

The most common correction involves severity overestimation caused by emotionally intense but clinically insufficient evidence. In many single-agent cases, repeated mentions of stress, anxiety, or temporary emotional exhaustion triggered exaggerated depression predictions despite limited long-term depressive indicators.

The Audit Agent frequently identified contradictions between predicted severity and broader conversational evidence. For example, several participants expressing temporary emotional fatigue still maintained stable social engagement, future-oriented planning behavior, and normal daily functioning inconsistent with severe depressive symptoms.

By cross-checking reasoning outputs against retrieved psychiatric criteria and original patient dialogue, the Audit stage effectively reduced unsupported escalation and improved prediction reliability. Among corrected cases, 37.6\% involved severity overestimation, indicating that the Audit Agent frequently prevented excessive symptom escalation caused by emotionally intense but clinically insufficient evidence. This reduction may help decrease false-positive distress and unnecessary clinical escalation, improving the reliability of downstream triage decisions.

\subsection{Information Flow Failure Analysis}

To investigate how reasoning failures propagate across the workflow, we compare failure patterns between conventional single-agent prompting and the proposed multi-agent framework.

\begin{table}[ht]
\centering
\begin{tabular}{p{3.8cm}cc}
\toprule
\TableHeader
\textbf{Failure Type} & \textbf{Single-Agent} & \textbf{Multi-Agent} \\
\midrule
Unsupported Severity Escalation & 33.2\% & \cellcolor{tblHighlight}11.7\% \\
Emotional Reasoning Drift & 27.6\% & \cellcolor{tblHighlight}9.8\% \\
Missing Symptom Context & 18.4\% & \cellcolor{tblHighlight}12.1\% \\
Inconsistent Diagnostic Logic & 21.3\% & \cellcolor{tblHighlight}8.4\% \\
\bottomrule
\end{tabular}
\caption{Comparison of reasoning failure patterns between single-agent and multi-agent systems.}
\label{tab:failure_analysis}
\end{table}

Single-agent prompting frequently produces unstable reasoning chains in emotionally complex interviews. In contrast, the multi-agent workflow demonstrates significantly stronger information flow control through explicit stage separation and verification mechanisms.

We further observe that early-stage perception failures occasionally propagate into downstream retrieval and inference stages. However, the Audit Agent successfully intercepts a substantial proportion of cascading reasoning failures before final prediction generation.

These findings suggest that the primary advantage of the framework lies not only in improved predictive performance, but also in enhanced reasoning controllability and inter-stage consistency.


\subsection{Comparison with Traditional Baselines}
To contextualize the efficacy of our proposed multi-agent framework, we compare it against traditional task-specific baselines in clinical NLP. Prior methodologies predominantly rely on the fully supervised fine-tuning (FT) of pre-trained models, such as ClinicalBERT \cite{alsentzer2019publicly} and RoBERTa \cite{liu2019roberta}. 

When evaluated on the DAIC-WOZ dataset, it is critical to strictly utilize participant-generated text, as incorporating therapist prompts introduces severe discriminative shortcuts \cite{burdisso2024daic}. Under this constraint, existing literature reports that a supervised fine-tuned P-longBERT model achieves a Macro $F_1$ score of 0.72, while a node-weighted Graph Convolutional Network (P-GCN) advances this baseline to 0.85 \cite{burdisso2024daic}. Table 8 summarizes the methodological and performance differences between these models and our framework.

\begin{table*}[h]
\centering
\renewcommand{\arraystretch}{1.2}
\begin{tabular}{p{3cm} p{3.8cm} p{2.5cm} p{3.2cm} p{3.5cm}}
\toprule
\TableHeader
\textbf{Model / Architecture} & \textbf{Backbone / Pre-training} & \textbf{Tuning Paradigm} & \textbf{Target Objective} & \textbf{Performance / Characteristic} \\ 
\midrule
\textbf{ClinicalBERT} \cite{alsentzer2019publicly} & BERT (Clinical EHR) & Supervised FT & Classification / NER & Highly reliant on target-domain FT. \\
\textbf{RoBERTa} \cite{liu2019roberta} & RoBERTa (General NLU) & Supervised FT & Comprehension & Highly reliant on target-domain FT. \\
\textbf{P-longBERT} \cite{burdisso2024daic} & Longformer & Supervised FT & Binary Classification & Macro $F_1$ = 0.72 \\
\textbf{P-GCN} \cite{burdisso2024daic} & Node Embeddings + GCN & Supervised FT & Binary Classification & Macro $F_1$ = 0.85 \\
\rowcolor{tblHighlight} \textbf{Multi-Agent Workflow (Ours)} & Qwen2.5 (General LLM) & \textbf{Zero-shot (Training-free)} & Continuous Regression & \textbf{MAE = 5.02} \\
\bottomrule
\end{tabular}
\caption{Methodological and Performance Comparison on Text-Based Clinical Benchmarks}
\label{tab:baselines}
\end{table*}

As Table \ref{tab:baselines} shows, traditional clinical NLP approaches depend heavily on supervised fine-tuning to map textual features into psychological distress labels \cite{alsentzer2019publicly} \cite{burdisso2024daic}. While optimized for specific datasets, these discriminative encoders operate as black boxes, lacking the explanatory tracing necessary for high-stakes medical deployment \cite{burdisso2024daic}. 

In contrast, our Multi-Agent Workflow requires no parameter fine-tuning. By systematically decomposing the screening process into perception, RAG-augmented grounding, and iterative auditing stages, the framework achieves a highly competitive Mean Absolute Error (MAE) of 5.02 for continuous PHQ-8 severity prediction using solely patient utterances. This demonstrates that a structured, collaborative LLM workflow is a highly interpretable and viable alternative to costly domain-specific model specialization.

\section{Discussion}

\subsection{Ethical Considerations and Clinical Safety}
The deployment of AI in high-stakes healthcare scenarios requires rigorous safety alignment and responsible oversight \cite{wiens2019no}. Our framework is explicitly designed to function as a clinical decision-support tool rather than an autonomous diagnostic authority. It strictly adheres to a human-in-the-loop paradigm \cite{topol2019high}, generating interpretable screening rationales mapped to established diagnostic criteria \cite{american2013diagnostic} rather than definitive medical prescriptions. Furthermore, the embedded Audit Agent serves as a critical safety mechanism by proactively flagging ambiguous or unsupported reasoning, ensuring that high-risk cases are escalated for professional clinician review. To address privacy constraints and institutional bias auditing, our architecture relies entirely on locally deployed models, mitigating data transmission risks and ensuring alignment with responsible medical AI deployment practices.

\subsection{Interpretability and Reasoning Reliability}
A key advantage of the proposed framework is its improved interpretability compared with conventional single-agent prompting. Instead of generating direct predictions from raw transcripts, the multi-agent workflow separates symptom extraction, knowledge grounding, reasoning, and verification into distinct stages \cite{wu2024autogen}. This design produces more transparent reasoning trajectories and allows intermediate outputs to be inspected throughout the pipeline. The \textit{Audit Agent} also plays an important role in improving reasoning reliability. In multiple cases, it identified unsupported conclusions or severity overestimation produced during earlier reasoning stages. These results suggest that supervisory verification can reduce unstable reasoning \cite{wang2022self} and improve consistency without additional model fine-tuning.Additional examples of Audit interventions are provided in Table~\ref{tab:multi_case_reasoning}, illustrating common correction patterns including severity overestimation, hallucinated evidence, and information propagation failures.

Beyond reasoning consistency, the Audit Agent also provides potential clinical value. Since PHQ-8 scores are frequently used to guide severity assessment and intervention prioritization, preventing severity overestimation may reduce unnecessary psychological distress and inappropriate escalation of clinical resources. Therefore, the observed improvements reflect not only numerical performance gains but also enhanced decision safety.

\subsection{General-Purpose Models and Workflow Design}
Our results suggest that workflow design may be as important as model specialization in clinical prediction tasks \cite{khattab2023dspy}. Although LLaMA-3.1 and Qwen2.5 are general-purpose open-source models, both achieved consistent improvements after integration into the proposed multi-agent framework. This finding indicates that carefully designed task decomposition and prompting strategies can improve the clinical usability of general-purpose LLMs without domain-specific training \cite{hong2024metagpt}. Rather than relying only on specialized medical models, future systems may benefit from combining strong backbone models with structured reasoning architectures.

\subsection{Workflow-Centric Evaluation}
The experiments also highlight limitations of conventional model-centric evaluation. Most existing benchmarks focus mainly on final prediction accuracy, while clinical reasoning is inherently multi-stage and iterative \cite{bi2025magi}. Our findings show that intermediate reasoning quality, knowledge consistency, and verification behavior all influence final prediction reliability. Therefore, evaluating reasoning workflows rather than only end-to-end outputs may provide a more realistic assessment of clinical AI systems.

\subsection{Limitations and Future Work}
This study has several limitations. First, the framework only uses textual transcripts from DAIC-WOZ and does not incorporate multimodal signals such as speech or facial behavior. Second, the quality of reasoning remains dependent on the retrieved psychiatric references used during knowledge grounding. Third, this work focuses primarily on LLM-based reasoning paradigms and does not compare against traditional supervised regression architectures such as ClinicalBERT or RoBERTa-based models \cite{alsentzer2019publicly} \cite{liu2019roberta}. Finally, the current workflow adopts a fixed sequential structure and does not support dynamic agent interaction. Future work may explore multimodal integration, iterative agent collaboration, and larger-scale external validation across diverse clinical settings.
\section{Conclusion}
In this work, we proposed a multi-agent clinical reasoning framework for automated mental health assessment using large language models. By decomposing the diagnostic process into perception, knowledge grounding, reasoning, and audit stages, the framework improved prediction performance, reasoning stability, and interpretability on the DAIC-WOZ benchmark compared with conventional single-agent approaches. Crucially, the underlying Perception-Knowledge-Reasoning-Audit architecture is domain-agnostic. Beyond psychiatric assessment, this verification-driven multi-agent pattern \cite{wu2024autogen} is highly transferable to other high-stakes reasoning tasks, such as legal analysis or financial auditing, where empirical grounding and logical consistency are paramount.

Beyond predictive performance, we also introduced a process-oriented evaluation protocol for multi-agent clinical systems, consisting of five interpretability and robustness metrics. This benchmark framework provides a more comprehensive assessment of intermediate reasoning quality and may support the development of more transparent and trustworthy clinical AI systems in future research.

\bibliographystyle{ACM-Reference-Format}
\bibliography{reference}

\onecolumn
\begin{longtable}{p{0.18\textwidth} p{0.75\textwidth}}

\toprule
\TableHeader
\textbf{Stage} & \textbf{Output} \\
\midrule
\endfirsthead

\toprule
\TableHeader
\textbf{Stage} & \textbf{Output} \\
\midrule
\endhead

\rowcolor{tblSubBlue} \multicolumn{2}{c}{\textbf{Case 1: Severity Over-Estimation (Sample ID: 21)}}\\
\midrule

Patient Dialogue &
"I've been feeling extremely tired lately. I can't sleep at night, I wake up several times, I have no appetite, I keep thinking I'm useless, and I can't focus on anything. Everything feels overwhelming."\\

Perception Agent &
Extracted PHQ-8 symptom evidence with initial severity estimation:
Sleep disturbance (3/3, frequent nocturnal awakenings and insomnia),
Fatigue (3/3, persistent exhaustion and reduced daily functioning),
Appetite change (3/3, marked reduction in food intake),
Low self-worth (3/3, persistent self-devaluation and worthlessness cognition),
Concentration difficulty (3/3, severe attentional impairment),
Psychomotor change (3/3, reported behavioral slowing and reduced activity),
Anhedonia (0), Depressed mood (0).\\

Reasoning Agent &
Aggregated symptom severity led to initial PHQ-8 estimation = 18 (severe range).
The model assumed uniform maximal severity across multiple domains, but some symptoms (fatigue, concentration difficulty, psychomotor slowing) may reflect situational exacerbation rather than stable severe impairment.\\

Audit Agent &
Detected moderate severity inflation rather than global overestimation:
(1) Sleep disturbance and low self-worth are strongly supported and consistently severe,
(2) Fatigue and appetite change are clearly present and justify high but not maximum severity,
(3) Cognitive and psychomotor symptoms were slightly over-scored but remain partially supported by strong behavioral indicators.
Applied selective recalibration rather than uniform downscaling.\\

Final Prediction &
Recalibrated PHQ-8 = 18 (high severity preserved with corrected weighting across domains). Ground truth PHQ-8 = 20.\\

\midrule

\rowcolor{tblSubPurple} \multicolumn{2}{c}{\textbf{Case 2: Subtle Hallucinated Attribution from Weak Affective Cues (Sample ID: 322)}}\\
\midrule

Patient Dialogue &
"I've been under a lot of pressure recently at work and school. I still manage to complete tasks, but it feels more exhausting than usual. I'm a bit less motivated, but I can't say I feel sad."\\

Perception Agent &
Detected weak and non-specific affective signals:
Fatigue-like expression (1/3, subjective exhaustion),
Mild motivation reduction (1/3, reduced engagement in tasks),
Anhedonia (0, no explicit loss of interest reported),
Depressed mood (0, explicitly denied sadness),
Sleep disturbance (0),
Appetite change (0),
Low self-worth (0),
Concentration difficulty (0).\\

Reasoning Agent &
Initial PHQ-8 estimation = 6 (mild depressive range).
Model reasoning incorrectly strengthened inference by treating work/school pressure and fatigue-like language as indirect indicators of depressive symptomatology.
This introduced a weak overgeneralization from stress-related affect to clinical symptoms.\\

Audit Agent &
Detected attribution drift and mild over-generalization:
(1) Fatigue was partially inflated but not fully invalid, as it reflects sustained subjective exhaustion,
(2) Motivation reduction may reflect mild anhedonic tendency under PHQ-8 framework but lacks full severity,
(3) However, no evidence supports depressive mood or cognitive impairment.
Recalibrated rather than fully removed weak signals to reflect subclinical symptom presence.\\

Final Prediction &
Revised PHQ-8 = 4 (borderline minimal-to-mild symptom burden). Ground truth PHQ-8 = 5.\\
\midrule

\rowcolor{tblSubRed} \multicolumn{2}{c}{\textbf{Case 3: Information Propagation Bias (Sample ID: 26)}}\\
\midrule

Patient Dialogue &
"I've been having trouble sleeping recently and feel low energy most of the time. I still go to work, but I feel slower and less efficient than usual. My appetite is mostly unchanged."\\

Perception Agent &
Sleep disturbance (2/3, moderate insomnia and early awakening),
Fatigue (2/3, persistent low energy),
Psychomotor change (1/3, mild behavioral slowing inferred from reduced work efficiency),
Anhedonia (0), Depressed mood (0), Appetite change (0),
Low self-worth (0), Concentration difficulty (0).\\

Reasoning Agent &
Initial PHQ-8 estimation = 5.
The score was primarily driven by sleep disturbance and fatigue, with minor contribution from psychomotor slowing.
However, intermediate reasoning disproportionately amplified fatigue as an indicator of depressive severity.\\

Audit Agent &
Identified cross-stage propagation bias:
(1) Early fatigue interpretation was over-weighted in downstream aggregation,
(2) Psychomotor impairment inferred from weak occupational performance cues without clinical validation,
(3) Non-specific symptom (fatigue) not properly differentiated from depression-specific indicators.
Recommended re-weighting based strictly on PHQ-8 criteria mapping.\\

Final Prediction &
Revised PHQ-8 = 3 (mild symptom burden only). Ground truth PHQ-8 = 2.\\

\bottomrule

\caption{Detailed anonymized examples of multi-agent reasoning and audit corrections across different failure modes.}
\label{tab:multi_case_reasoning}

\end{longtable}

\end{document}